\documentclass[acmsmall]{acmart}
\AtBeginDocument{%
  \providecommand\BibTeX{{%
    \normalfont B\kern-0.5em{\scshape i\kern-0.25em b}\kern-0.8em\TeX}}}

\usepackage{graphicx}
\usepackage{multirow}
\begin{document}

\title{LabelPrompt: Effective Prompt-based Learning for Relation Classification}

\author{Wenjie Zhang}
\email{zhangwenjie@stu.jiangnan.edu.cn}
\orcid{0000-0002-3307-6189}
\affiliation{%
  \institution{Jiangnan University}
  \streetaddress{1800 Lihu Avenue}
  \city{Wuxi}
  \state{Jiangsu}
  \country{China}
}

\author{Xiaoning Song}
\email{x.song@jiangnan.edu.cn}
\authornote{Corresponding author}
\affiliation{%
  \institution{Jiangnan University}
  \streetaddress{1800 Lihu Avenue}
  \city{WuXi}
  \state{JiangSu}
  \country{China}
}

\author{Zhenhua Feng}
\email{z.feng@surrey.ac.uk}
\affiliation{%
  \institution{University of Surrey}
  \streetaddress{Guildford GU2 7XH}
  \city{Surrey}
  \country{United Kingdom}
}

\author{Tianyang Xu}
\email{tianyang.xu@jiangnan.edu.cn}
\affiliation{%
  \institution{Jiangnan University}
  \streetaddress{1800 Lihu Avenue}
  \city{WuXi}
  \state{JiangSu}
  \country{China}
}

\author{Xiaojun Wu}
\email{wu_xiaojun@jiangnan.edu.cn}
\affiliation{%
  \institution{Jiangnan University}
  \streetaddress{1800 Lihu Avenue}
  \city{WuXi}
  \state{JiangSu}
  \country{China}
}
\renewcommand{\shortauthors}{Zhang, et al.}

\begin{abstract}
    Recently, prompt-based learning has gained popularity across many natural language processing (NLP) tasks by reformulating them into a cloze-style format to better align pre-trained language models (PLMs) with downstream tasks. However, applying this approach to relation classification poses unique challenges. Specifically, associating natural language words that fill the masked token with semantic relation labels (\textit{e.g.} \textit{``org:founded\_by}'') is difficult. To address this challenge, this paper presents a novel prompt-based learning method, namely LabelPrompt, for the relation classification task. Motivated by the intuition to ``GIVE MODEL CHOICES!'', we first define additional tokens to represent relation labels, which regard these tokens as the verbaliser with semantic initialisation and explicitly construct them with a prompt template method. Then, to mitigate inconsistency between predicted relations and given entities, we implement an entity-aware module with contrastive learning. Last, we conduct an attention query strategy within the self-attention layer to differentiates prompt tokens and sequence tokens. Together, these strategies enhance the adaptability of prompt-based learning, especially when only small labelled datasets is available. Comprehensive experiments on benchmark datasets demonstrate the superiority of our method, particularly in the few-shot scenario.
\end{abstract}





\maketitle

\section{Introduction}\label{sec:introduction}

Pretrained Language Models (PLMs) have achieved great success in many Natural Language Processing (NLP) tasks~\cite{jiang-etal-2020-smart,xu-etal-2021-bert,guo-etal-2022-longt5,zhong-chen-2021-frustratingly}, such as Relation Classification (RC), Named Entity Recognition (NER), etc. The strong performance of PLMs stems from their ability to learn rich knowledge representations from large unlabelled corpora through self-supervised pretraining objectives such as Masked Language Modelling (MLM). A common paradigm is to then fine-tune the pretrained parameters on downstream task datasets, thereby transferring the embedded knowledge in the PLM to improve performance on the target task~\cite{Peters2018DeepCW, Lewis2020BARTDS}.
However, a fundamental challenge with this paradigm is the inherent mismatch between the pretraining objective of the PLM and the actual downstream task. This discrepancy in objectives can result in a sub-optimal transfer of knowledge from pretraining to fine-tuning~\cite{liu2021pre}.
% This necessitates bridging the gap between them when applying PLMs to a specific task for effective model training and language knowledge preserving of the pretrained language models.

To bridge the mismatch between pretraining and downstream tasks, prompt-based learning  has emerged as an effective approach for utilising PLMs~\cite{lester-etal-2021-power,brown2020language,wallace-etal-2019-universal}.
The key idea is to reformulate a downstream task into the masked language modelling format used during pretraining. This is achieved through two main components: 1) a prompt template that prepends instructions and context to the input, and 2) an answer mapping method (known as a verbaliser) that converts the model's output into target labels
By transforming tasks into the native pretraining format, prompt-based learning reduces the discrepancy in objectives between pretraining and fine-tuning~\cite{tan-etal-2022-msp,ma-etal-2022-template}.
Despite promising results, manually engineering prompt templates and verbalisers can be expensive and time consuming, limiting wider adoption~\cite{liu2021pre}.
Recent work has focused on automatically generating prompts~\cite{lester-etal-2021-power,liu-etal-2022-p} or learning continuous prompt representations~\cite{shin-etal-2020-autoprompt,li-liang-2021-prefix} to reduce the cost of prompt engineering. Overall, prompt-based learning offers an appealing paradigm for effectively utilising PLMs, warranting further research to address current limitations.

Meanwhile, inspired by the great success of prompt-based learning, recent studies have explored applications to text classification tasks~\cite{zhang-etal-2022-prompt-based}.
In this paper, we specifically investigate prompts for relation classification, a prevalent text classification problem.
Given a pair of entities mentioned in a text, the goal of relation classification is to predict the semantic relation between them, if any (\textit{e.g.}, ``\textit{per:employee\_of}'', ``\textit{org:founded}'' and ``\textit{no\_relation}'')~\cite{hendrickx-etal-2010-semeval}.
While prompt-based learning provides an appealing approach to use pretrained language models for this task, applying prompts to relation classification also poses unique challenges.

First, relation classification presents a unique challenge compared to many NLP tasks in that the target labels contain rich semantic information, as opposed to being single words representable within the pretrained model's vocabulary. Consequently, existing answer mapping techniques struggle to effectively project the model's masked outputs to the complex, multi-word relation labels when reformulating relation classification as a masked language modelling task.

To improve mapping performance, some studies suggest searching external data for label-related words to design a verbaliser~\cite{hu-etal-2022-knowledgeable, gu-etal-2022-ppt,zhu2022prompt}. However, these methods have yet to achieve optimal results.
Alternatively, other approaches define additional learnable tokens external to the pretrained vocabulary, and map model outputs to these new tokens for classification~\cite{chen2022knowprompt}.
Although these methods can mitigate token mapping issues,  tuning the new tokens requires more labelled training data due to the significant gap from the pretrained vocabulary. So the first challenge in relation extraction is developing effective training methodologies that can enhance model performance under limited data availability.

Second, the masked language modelling (MLM) pre-training objective depends on predictions derived from neighbouring words and global sentence features. However, relation classification requires not only identifying a relation from a sentence, but also determining whether the relation holds between two given entities. Existing approaches predominantly predict relations at the sentence level, without explicitly modelling the correspondence between the entity-relation triplet $(s, r, o)$.

For instance, consider an abbreviated excerpt from the ReTACRED~\cite{retacred} dataset: ``\textit{Travis the Chimp, the 14-year-old, 200-pound pet of Sandra Herold, 71.}'' In this example, ``\textit{Sandra Herold}'' is denoted as the subject entity while ``\textit{14-year-old}'' is denoted as the object entity. Despite the ground truth relation existing between ``\textit{Sandra Herold}'' and ``\textit{71}'', the model incorrectly predicts the relation as ``\textit{per:age}'' between ``\textit{Sandra Herold}'' and ``\textit{14-year-old}''. Evidently, the model fails to accurately capture the correspondence between the given entities and their relation. Subsequently, an additional challenge arising in relation classification is augmenting the model's capacity to discern given entities when inferring relations.

To mitigate the above limitations of prompt-tuning for relation classification, we develop a novel LabelPrompt approach with an entity-aware module.
LabelPrompt is a simple yet effective method that significantly reduces the gap between pre-training objectives and the relation classification task. Unlike previous methods, LabelPrompt enables the model to capture relation labels more intuitively by augmenting the input sequence.

In terms of the studies in prompt templates, many researchers have focused on constructing a prompt template that meet the requirements of the task formulation.
At the same time, some studies have shown that the effectiveness of the prompt template has a significant impact on task performance.
Thus, we believe it is essential to provide the pre-trained model with supplementary task-relevant information via the input.
Building on this idea, we suggest an intuitive technique for providing the model with selectable options. In particular, we develop a prompt template approach that embeds relation class labels into the input example before the model is ingested.

Specifically, to enhance the ability of the masked language modelling (MLM) for relation label prediction, we propose extending the vocabulary space with custom label tokens that are not present in natural language (\textit{e.g.}, ``\textit{[C1]}''). The embeddings of these tokens are initiated with semantic information derived from relation label texts. However, this introduces a gap between pre-training and our task, as the additional label tokens are also not present in the pre-training vocabulary space. To mitigate this limitation and improve prompt-based method performance in few-shot scenarios,  we expand the vocabulary with additional label prompt tokens and use these label tokens to construct prompt templates on the input side. Furthermore, to prevent additional prompt tokens from affecting sentence semantics, we redesign the attention query strategy that use distinct query projections for prompt and sentence token pairs. This controls prompt-sentence interactions, and allows label prompt tokens to provide task-specific information without compromising sentence semantics encoded in the pre-trained language model. This promotes independence between prompts and sentences to improve the efficacy of relation classification.

Second, to enhance entity perception during model training, we implement an entity-aware module to assess the correlation between entities and their relations. As shown in Fig.~\ref{fig:main}, we adopt a contrastive learning-inspired approach to construct positive and negative samples, where the former consist of given entities and their predicted relation, while the latter contains randomly selected tokens and ground-truth relations within the sentence. This effectively constrains the relation and entities with semantic information.

Finally, extensive evaluations on popular benchmark datasets demonstrate that our proposed approach yields significant performance improvements compared to fine-tuning and other prompt-based methods. We attribute this enhancement to our method's ability to effectively constrict the model's search space and guide optimisation towards the correct direction.

To summarise, the main contributions of the proposed LabelPrompt method include:
\begin{itemize}
    \item We introduce a prompt-based method that explicitly exploits relation features and significantly improves the performance of prompt-based learning in few-shot scenarios.
    \item We implement an entity-aware module that employs contrastive learning to enhance entity awareness during the inference stage.
    \item We design an attention query strategy within self-attention layers to differentiate between prompt and sentence tokens and improve the performance of prompt-based learning.
\end{itemize}

The rest of this paper is organised as follows. We first introduce the related work in Section~\ref{sec:relat}. Then we present the proposed LabelPrompt method in Section~\ref{sec:method}. The experimental results and analysis of the proposed method are reported in Section~\ref{sec:experiment} and Section~\ref{sec:analysis}, respectively. Last, we draw the conclusion of the proposed method in Section~\ref{sec:conclusion}.

\section{Related Work}
\label{sec:relat}
\subsection{Pre-training and Fine-tuning}
Pre-trained language models are trained in large-scale unlabelled text data to learn robust general-purpose features, such as lexical, syntactic, semantic, and word representation~\cite{liu-etal-2022-p}. Many big models, such as GPT~\cite{liu2021gpt}, BERT~\cite{devlin2019bert}, RoBERTa~\cite{liu2019roberta}, T5~\cite{raffel2020exploring}, have been proposed, greatly promoting the development of the research in natural language processing.
Most NLP tasks benefit from the knowledge contained in the pre-training language model, and the increase in the size of a pre-training model as well as data facilitates the task.
% The powerful textual representation and knowledge of the pre-trained language model are useful for most NLP tasks, and the advent of large models has greatly improved their representation capability.
By fine-tuning the parameters of PLMs with additional specific modules and task-specific objective functions, the models can be adapted to most downstream tasks.
Generally, the pre-training and fine-tuning paradigm become the foundation for many NLP tasks, such as named entity recognition~\cite{wu-etal-2021-mect}, relation classification~\cite{lyu-chen-2021-relation}, and question answering~\cite{2022ProQA}, etc.

\subsection{Prompt-based Learning}
Although fine-tuning has achieved significant success in many fields, there remains a large gap between the objectives of pre-training and fine-tuning tasks~\cite{liu2021pre}. Typically, during pre-training, a PLM is trained using self-supervised corpus texts with Masked Language Modelling (MLM) and Next Sentence Prediction (NSP) tasks~\cite{devlin2019bert}. This differs from other NLP tasks, such as classification and sequence tagging tasks. As a result, recent studies have sought to explore how to effectively use pre-trained language models and address this gap~\cite{wang-etal-2021-k, moiseev-etal-2022-skill}.

Prompt-based learning is a different approach that aims to reduce the gap between pre-training and fine-tuning. Rather than fine-tuning, prompt-based learning reformulates a downstream task to the original PLM training task. For example, a prompt template might transfer a machine translation instance as ``\textit{English: I love this film. French: \_\_}''. A large PLM could then fill in the blank with a French translation of the English text. Essentially, prompt-based learning enables PLMs to perform downstream tasks by providing additional information in the form of a prompt~\cite{lester-etal-2021-power,gao-etal-2021-making}.

Many studies have shown that prompt templates containing semantics can effectively activate knowledge in the model. Building on this idea, \citet{2019Language} considered language models as knowledge bases and proposed the LAMA (LAnguage Model Analysis) probe dataset. The LAMA probe provides manually created completion templates for exploring the knowledge in PLMs. By using hand-crafted query sentences, it extracts the knowledge base from PLMs and performs comparably to using PLMs directly. However, the manual design of templates as discrete prompts is time-consuming, expensive~\cite{shin-etal-2021-constrained}, and often requires a wealth of expert knowledge.

To extend prompt-based learning to a wider range of tasks and reduce the overhead of prompt text construction, \citet{2020AutoPrompt} automatically generated prompts for different task sets based on a gradient search approach. Some researchers have also added a small set of task-specific prefixes to replace manual templates as continuous prompts, which are learnable by backpropagation~\cite{li-liang-2021-prefix, lester-etal-2021-power}. Additionally, \citet{zhong-etal-2021-factual} suggests initialising continuous prompts based on the discrete prompts already explored by prompt search methods.

Another important component of prompt-based learning is the answer-mapping approach. Since the masked output does not always match the label, the aim of answer mapping methods is to map the model output to the answer space when the answers cannot be considered as task outputs. For example, sentiment analysis tasks require mapping the output of the model template ``\{ \textit{good, fantastic, unhappy, average} \}'' to ``\{ \textit{positive, negative, neutral} \}''.
The answer-mapping method is also known as a verbaliser. A verbaliser is a projection method that constructs a project between a natural word space and a task label space~\cite{hu-etal-2022-knowledgeable}. Just like the prompt template, a verbaliser also has ``manual design''~\cite{yin-etal-2019-benchmarking, cui-etal-2021-template}, ``discrete answer search''~\cite{jiang-etal-2020-know, 9914670} and ``continuous answer search'' strategies~\cite{hambardzumyan-etal-2021-warp}. Equally, the quality of a verbaliser can have an impressive impact on model performance.

In conclusion, prompt-based learning is widely considered as a highly effective method for utilising pre-trained language models, and it is a subject worthy of further study.

\begin{table}
\center
\caption{Two examples of relation classification. Relation classification aims to predict the relation between two entities mentioned in a given sentence.}
\label{tab:example}
\begin{tabular}{p{0.65\linewidth}|p{0.25\linewidth}}
\toprule
\textbf{Sentence} & \textbf{Relation}  \\
\midrule
\textit{Mark Fisher [subject]} writes for the \textit{Dayton Daily News [object]}. &  per:employee\_of \\
\midrule
He has a sense of \textit{humor [subject]} about his \textit{reaction [object]} to that day. & no\_relation \\
\bottomrule
\end{tabular}
\end{table}

\subsection{Relation Classification}
Relation Classification (RC) is a sub-task of Information Extraction (IE).
The goal of RC is to identify the semantic relation between two entities in a given text. As shown in Table~\ref{tab:example}, given the sentence ``\textit{Mark Fisher writes for the Dayton Daily News}.'', the relation between the subject ``\textit{Mark Fisher}'' and the object ``\textit{Dayton Daily News}'' is ``\textit{per:employee\_of}''. RC is a crucial task for many natural language understanding applications, including question answering, knowledge base construction, and text summarization~\cite{hendrickx-etal-2010-semeval}.

Early approaches for RC relied heavily on prior knowledge to extract additional features from the text~\cite{califf-mooney-1997-relational, zeng-etal-2015-distant}. The emergence of large-scale PLMs, such as BERT~\cite{devlin2019bert}, has provided universal language representations that are useful for RC tasks~\cite{NIPS2017_3f5ee243, Cohen2020RelationEA, yamada-etal-2020-luke}. However, these methods often require significant amounts of labelled data and complex, task-specific neural modules, and tend to underperform in the few-shot scenario.

Recently, there have been studies exploring the use of prompt-based learning to leverage the knowledge contained in PLMs for relation classification. The LAMA dataset is a probe for analysing factual and commonsense knowledge in PLMs, which consists of a set of knowledge sources, each comprising a set of facts. For example, one template could be "\textit{\_\_ is the capital of France.}" \citet{hu-etal-2022-knowledgeable} proposed a method that uses external knowledge to construct a verbaliser for obtaining more accurate label words for text classification. \citet{han2021ptr} applied logic rules to construct prompts with several sub-prompts, effectively using prior knowledge from relation classification. \citet{chen2022knowprompt} proposed a prompt-based learning approach called KnowPrompt that incorporates abundant semantics and prior knowledge in relation labels into relation classification.

All these methods aimed to effectively use prompt-based learning to narrow the gap between relation classification and pre-training tasks, and demonstrated promising results in both few-shot and full-data scenarios. However, none of these methods addressed the issue of inconsistency between prompt output and relation label texts. Inspired by KnowPrompt, we propose a new prompt-based learning approach, namely LabelPrompt, that explicitly constructs prompt templates using label tokens and adds supervised information corresponding to the labels at the model's output.

\section{The Proposed LabelPrompt Method}\label{sec:method}
\begin{figure*}[t]
    \centering
    \includegraphics[width = 1.0\linewidth]{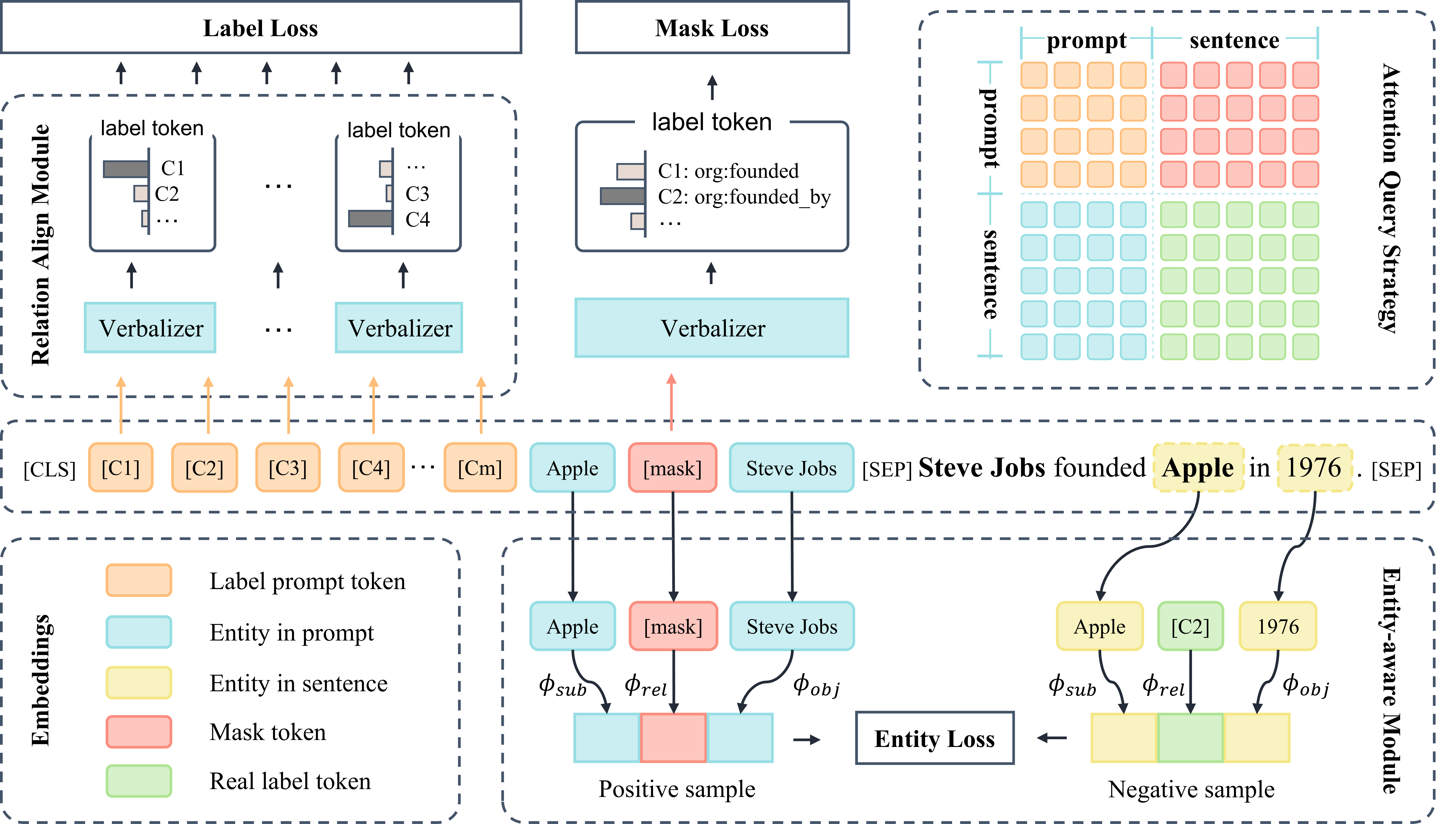}
    \caption{An overview of the proposed method. We input the tokens, including the label tokens, into the RoBERTa model to obtain a global feature representation and improve contextual awareness. The masked token is then used to predict the relation of the sentence. The relation-align module aligns the encoded label prompt tokens with their corresponding classes and computes the label loss to preserve the label prompt information. The Entity-Aware Module constructs two samples based on the triplet information and the sentence tokens in the pure prompt template to constrain the correspondence between the given entities. The attention query strategy in the encoding stage facilitates independent lines of prompt tokens and sentence tokens.}
    \label{fig:main}
\end{figure*}

Relation Classification (RC) can be denoted as $\mathcal{T} = \{ \mathcal{X}, \mathcal{Y} \}$, where $\mathcal{X}$ is the set of instances and $\mathcal{Y}$ is the set of relation labels. For each instance $\mathbf{x} \in \mathcal{X}$, it contains a sentence  $x = \{x_1, x_2, \dots x_n\}$, two entities ($e_s, e_o$) mentioned in $x$ , and its label relation $y_x \in \mathcal{Y}$.
The goal of relation classification is to predict the relation $y \in \mathcal{Y}$ between given subject entity $e_s$ and object entity $e_o$ in sentence $x$. If there is no relation between the entities, it will be considered as a special type of relation, \textit{i.e.} ``\textit{no\_relation}'', which exists in the set of relations.

We will elaborate on the major components and training schemes of our approach in the following parts. As illustrated in Fig.~\ref{fig:main}, the input to the model is composed of two primary components, the prompt tokens and the sentence tokens. We can further separate the prompt tokens into label prompt tokens and typical prompt tokens.

\subsection{Label Prompt}

\subsubsection{Label Prompt Tokens}

For each relation, $y \in \mathcal{Y}$, its label text is not an individual word in natural language, but a composite text, \textit{e.g.} ``\textit{org:top\_members/employees}'', ``\textit{per:country\_of\_death}'' and etc. We define a label space by creating $m$ label tokens $\mathcal{C} = \{ c_1, c_2, \dots, c_m \}$ and extending it to the vocabulary of PLM, where $m$ represents the size of the relation set $\mathcal{Y}$. For better training setup, these label tokens will be initialised by their knowledge-rich semantic label texts. The initialisation process is defined as follows:

\begin{equation}
C_i = \frac{1}{N_i}\sum_{j=1}^{N_{i}}\textrm{Embedding}_{\mathcal{M}}(t_{j}),
\end{equation}
where $C_i$ represents the word embedding of label $c_i$, $\textrm{Embedding}_{\mathcal{M}}$ stands for the word embedding layer of $\mathcal{M}$, and $t_j$ is denoted as the $j$-th sub-text of the decomposed relation label text $T_i = \{ t_1, t_2, \dots , t_{N_i} \}$.

\subsubsection{Label Prompt Template}
To provide context for our work, we first introduce the concepts of prompt templates and the prompt template method. In this paper, a prompt template refers to additional text consisting of prompt tokens, while the template method refers to the contract function used to construct the input instance  $\mathbf{x}_{prompt}$ with a specific token $\texttt{[MASK]}$.

The objective of our work is to address the challenge of the prompt-based method to fit correct label tokens in the few-shot scenario where data is limited. To achieve this, we propose a solution that involves converting the relation classification task into a masked language modelling one. However, previous studies have shown that this approach may not be effective due to the large size of the vocabulary and the newly added label tokens that were not present during the model's pre-training stage.

To overcome this limitation, we introduce an intuitive approach of directly adding label tokens to the input sequences. Our proposed prompt template, namely Label Prompt Template, consists of both label prompt tokens and typical prompt tokens. By incorporating the label tokens directly into the input sequences, we can improve the performance of the prompt-based model in the few-shot scenario.  We represent the prompt template as $T(\mathbf{x})$:
\begin{equation}
T(\mathbf{x}) = \{ c_1, c_2, \dots, c_m,[\textrm{SEP}], e_s, [\textrm{MASK}], e_o \},
\end{equation}
where $c_i \in \mathcal{C}$ is the label token, $e_s$ and $e_o$ are the subject and object entities. $[\texttt{CLS}]$, $[\texttt{SEP}]$ and $[\texttt{MASK}]$ are predefined in $\mathcal{M}$. The input $x$ will be converted to the input sequence as follows,

\begin{equation}
x_{prompt} = \left\{[\textrm{CLS}],  T(\mathbf{x}), [\textrm{SEP}], x, [\textrm{SEP}] \right\}.
\label{equ:x-prompt}
\end{equation}

\subsubsection{Label Prompt Learning}

In this section, we introduce our key training objective. First, we define $\mathbf{h}$ as the output of the encoder of the model $\mathcal{M}$. An input sequence $x_{prompt}$ is encoded by the model to obtain $\mathbf{h} \in \mathbb{R}^{\ell \times d}$,

\begin{equation}
\mathbf{h} = \textrm{Encoder}_{\mathcal{M}}(x_{prompt}),
\end{equation}
where $\ell$ represents the length of $x_{prompt}$, and $d$ is the size of the hidden vectors.

We also define $h_{mask}$ as the hidden vector of the $[\texttt{MASK}]$, $h_{c_i}$ as the hidden vector of label prompt token $c_i$, $h_{sub}$ and $h_{obj}$ represent the hidden vectors of subject and object entities.

Next, we define a verbaliser $V_{\phi}$ as an answer-mapping method that maps a hidden vector $h$ to a relation label,

\begin{equation}
V_{\phi}(h) = \mathbf{E}_{label} \cdot  (W_{v} \cdot h + b),
\end{equation}
where $\mathbf{E}_{label}$ represents the labels' embedding, $W_{v}$ and $b$ are learnable parameters. We use $C_i \in \mathbf{E}_{label}$ to denote the embedding of label token $c_i$. Then we calculate the probability distribution of the mapping result of hidden vector $h$. The probability that the label of the masked token is $c_i$ is calculated using the \textit{softmax} function,

\begin{equation}
p\left(V_{\phi}\left(h\right) = c_i | x_{prompt}\right) = \frac{\exp(C_{i}  \cdot (W_{v} \cdot h + b))}{\sum_{j=1}^{m} \exp(C_{j} \cdot (W_{v} \cdot h + b))  },
\end{equation}
where $m$ is the size of the relation set $\mathcal{Y}$.

Last, our model optimisation aims to assign a suitable relation label to the masked token. We define the loss function,

\begin{equation}
\mathcal{L}_{mask} = \frac{1}{m} \sum_{i=1}^{m} y_x \log p(y_x|\mathbf{x}),
\end{equation}
as the cross-entropy loss between the predicted probability $p(y_x|\mathbf{x}) = p(V_{\phi}(h_{mask}) = \mathcal{C} | x_{prompt})$ and the ground truth $y$.
The label token with the highest probability will be selected as the predicted relation class between the entities in instance $\textbf{x}$.

\subsection{Relation-Align Module}
The objective of the relation-align module is to align the label tokens with the deep decision-making layer of the model. In the context of relation extraction models, the label token on the input side may lose its inherent features in the subsequent encoder layers, which reduces its effectiveness in aiding the prediction of relations in the deep decision-making layer.

To tackle this problem, we propose the relation-align module, which is integrated at the end of our model. Furthermore, the verbaliser $V_{\phi}$ is incorporated behind the hidden vector $h_{c_{i}}$ of each label token $c_i$ in $x_{prompt}$ to reinforce its semantic meaning, just like the mask loss. To accomplish this, the probability score $p(c_i|\mathbf{x}, c_i) = p\left(\tilde{c_{i}} = \mathcal{C} | x_{prompt}\right)$ is formulated for each label token, which can be classified into itself by the verbaliser, where $\tilde{c_{i}}$ denotes $V_{\phi}(h_{c_i})$.

The cross-entropy loss for each label token is computed based on the probability distribution, and then the losses are averaged to obtain the label loss $\mathcal{L}_{label}$:
\begin{equation}
\begin{aligned}
\mathcal{L}_{c_i} = \frac{1}{m} \sum_{i=1}^{m} & y_x \log p(c_i|\mathbf{x}, c_i), \\
\mathcal{L}_{label} =& \frac{1}{m }\sum_{i=1}^{m} \mathcal{L}_{c_i} .
 \end{aligned}
\end{equation}

\subsection{Entity-Aware Module}
In this section, we introduce an entity-aware module aimed at addressing the issue of a PLM failing to consider given entities when predicting the relation between them.

Through analysis of numerous error cases, we observed that a PLM can detect a relation in a sentence, but it may not exist between the two given entities. In other words, the model does not take the given entities into account when predicting the relationship.

Drawing from TransE~\cite{TransE-NIPS2013_1cecc7a7}, which suggests a correlation, $s + r = o$, between the features of entities ($s,o$) and the relation $r$, we propose the entity-aware module. The module employs the distance metric $d(s, r, o) = \lVert s + r - o\rVert_{2}$ to measure the distance between entities and relations.
Furthermore, the dimensions of the encoder outputs for entities and relations are reduced to eliminate redundant features:
\begin{equation}
\begin{aligned}
s &= \phi_{sub} \cdot h_{sub}, \\
o &= \phi_{obj} \cdot h_{obj}, \\
r &= \phi_{rel} \cdot h_{mask},
\end{aligned}
\label{equ:reduction}
\end{equation}
where $\phi_{sub}$, $\phi_{obj}$ and $\phi_{rel} \in \mathbb{R}^{p \times d}$ are trainable parameters, $p$ is the length of the hidden vector after dimension reduction.

In Equ.~(\ref{equ:reduction}), the tuple ($s, r, o$) is considered a positive example while negative examples ($s', r, o'$) are randomly sampled for the entity-aware loss, where $s'$ and $o'$ denote two arbitrary spans from the sentence. The entity-aware loss, denoted by $\mathcal{L}_{entity}$, is formulated as follows:

\begin{equation}
\begin{aligned}
\mathcal{L}_{entity} = &-\log \sigma\left(\gamma -d\left(s, r, o\right)\right)  \\ &- \log \sigma\left(d\left(s', r, o'\right) - \gamma\right),
\end{aligned}
\end{equation}
where $\sigma$ is the sigmoid function and $\gamma$ is the margin that is set to 0.3.

Different from previous approaches that use simple negative samples, we use more challenging negative samples and concentrate on the contrastive loss between current sentences during the training phase. Our experiments reveal that other samples in the same batch can hinder the convergence of this portion of the loss, thus we disregard them.

\subsection{Attention Query Strategy}

As shown in Equ.~(\ref{equ:x-prompt}), the input sequence comprises two distinct categories of tokens, namely prompt tokens and sentence tokens. However, the inclusion of additional unknown tokens can significantly impact the semantic meaning of the sentence and subsequently, the efficacy of relation extraction.

To mitigate the adverse impact of prompt tokens on sentence semantics, we employ an attention query strategy during the PLM encoding process. This approach utilises distinct query matrices for different token pairs in the self-attention layer to minimise the influence of prompt tokens on sentence semantics and enhance the quality of relation extraction outcomes. Further details are provided in Fig.~\ref{fig:main}.

The self-attention layer in the transformer encoder can be formulated as follow:

\begin{equation}
    \operatorname{SelfAttn}(E) = \operatorname{softmax}(\frac{QE \cdot (KE)^{T}}{\sqrt{d}}) \cdot VE,
\end{equation}
where $E$ represents the input and $Q \in \mathbb{R}^{d \times d}$, $K \in \mathbb{R}^{d \times d}$, $V \in \mathbb{R}^{d \times d}$ denote the mapping matrices. In our method, we select different query mapping matrices depending on which pair of tokens are interacting: $Q_{pp}$, $Q_{ps}$, $Q_{sp}$, and $Q_{ss}$. The effective initialisation of pre-training parameters enables the method to maintain strong few-shot performance even when additional query matrices is incorporated.

\subsection{Objective Function}

We propose three loss functions to optimise the model parameters: mask loss for relation prediction, label loss for semantic consistency of label prompt tokens, and entity loss for improving entity awareness.

Our objective is to minimise the final loss $\mathcal{L}$ which is the weighted sum of those losses:
\begin{equation}
\mathcal{L} = \mathcal{L}_{pred}\space  + \alpha_{1} \mathcal{L}_{label} \space + \alpha_{2} \mathcal{L}_{entity},
\end{equation}
where $\alpha_{1}, \alpha_{2}$ are the weights for the label loss and entity loss, respectively. In this paper, we set them to 1 and 0.04 based on our empirical experimental results.

\begin{table}
\center
\caption{Statistics of different datasets for relation classification.}
\label{tab:datasize}
\begin{tabular}{l|r|r|r|r}
\toprule
    Dataset     & \#Training   & \#Validation     & \#Test   & \#Relation \\
    \midrule
    SemEval     &6,507      &1,493      &2,717      &19 \\
    TACRED      &68,124     &22,631     &15,509     &42 \\
    TACREV      &68,124     &22,631     &15,509     &42 \\
    ReTACRED    &58,465     &19,584     &13,418     &40 \\
    \bottomrule
\end{tabular}
\end{table}

\section{Experiment}
\label{sec:experiment}
In this section, we evaluate the proposed method on four benchmark datasets with the widely used micro $F_1$ metric. When it comes to multi-class classification tasks, there are two commonly used averaging methods, namely macro and micro scores. The macro $F_1$ score calculates the global score by averaging over all the classes, while the micro-averaging method computes the global score by taking into account the True Positives (TP), False Negatives (FN), and False Positives (FP). In this paper, we use the micro $F_1$ score as the primary metric to evaluate these methods for all the experiments.

\begin{table*}[t]
\center
\caption{$F_1$ scores (\%) on the TACRED, TACREV, ReTACRED and SemEval in the few-shot scenario. The best results are marked \textbf{bold}. }
\label{tab:few-shot}
\begin{tabular}{l|l|c|c|c|c}
\toprule
$k$-shot                  & Methods                       & TACRED        & TACREV        & ReTACRED      & SemEval \\
\midrule
\multirow{5}{*}{8-shot} & \textsc{fine-tuning}          & 12.2          & 13.5          & 28.5          & 41.3    \\
& \textsc{GDPNet}~\cite{Xue2021GDPNetRL}                 & 11.8          & 12.3          & 29.0          & 42.0    \\
& \textsc{PTR}~\cite{han2021ptr}                         & 28.1          & 28.7          & 51.5          & 70.5    \\
& \textsc{KnowPrompt}~\cite{chen2022knowprompt}          & 32.0          & 32.1          & 55.3          & 74.3    \\
& \textsc{\textbf{LabelPrompt}}(ours)                   & \textbf{33.9} & \textbf{34.8} & \textbf{61.5} & \textbf{77.0}       \\
\midrule
\multirow{5}{*}{16-shot}& \textsc{fine-tuning}          & 21.5          & 22.3          & 49.5          & 65.2    \\
                        & \textsc{GDPNet}               & 22.5          & 23.8          & 50.0          & 67.5    \\
                        & \textsc{PTR}                  & 30.7          & 31.4          & 56.2          & 81.3    \\
                        & \textsc{KnowPrompt}           & \textbf{35.4} & 33.1          & 63.3          & \textbf{82.9}    \\
            & \textsc{\textbf{LabelPrompt}}(ours)       & 34.4          & \textbf{35.4} & \textbf{64.5} & 81.7       \\
\midrule
\multirow{5}{*}{32-shot}& \textsc{fine-tuning}          & 28.0          & 28.2          & 56.0          & 80.1    \\
                        & \textsc{GDPNet}               & 28.8          & 29.1          & 56.5          & 81.2    \\
                        & \textsc{PTR}                  & 32.1          & 32.4          & 62.1          & 84.2    \\
                        & \textsc{KnowPrompt}           & \textbf{36.5} & 34.7          & 65.0          & \textbf{84.8}    \\
            & \textsc{\textbf{LabelPrompt}}(ours)       & 35.4          & \textbf{36.8} & \textbf{66.7} & 84.6       \\
\bottomrule
\end{tabular}
\end{table*}

\begin{table*}[t]
\center
\caption{$F_1$ scores (\%) on the TACRED, TACREV, ReTACRED and SemEval in the full-data scenario. In the ``Extra Data'' column, ``w/o'' means that no additional data is used for pre-training and fine-tuning, yet ``w/'' means that extra data or knowledge bases are used for data augmentation. The best results are marked \textbf{bold}.}
\label{tab:all}
\begin{tabular}{l|c|c|c|c|c}
\toprule
Model                                                   & Extra Data & TACRED  & TACREV  & ReTACRED  & SemEval \\
\midrule
\multicolumn{6}{c}{Fine-tuning pre-trained models}               \\
\midrule
\textsc{RoBERTa\_large}~\cite{liu2019roberta}            & w/o   & 68.7   & 76.0   & 84.9      & 87.6    \\
\textsc{KnowBERT}~\cite{chen2022knowprompt}              & w/    & 71.5   & 79.3   & -         & 89.1    \\
\textsc{MTB}~\cite{baldini-soares-etal-2019-matching}    & w/    & 70.1   & -      & -         & 89.5    \\
\textsc{SpanBERT}~\cite{Joshi2020SpanBERTIP}             & w/    & 70.8   & 78.0   & 85.3      & -       \\
\textsc{LUKE}~\cite{yamada-etal-2020-luke}               & w/    & 72.7   & 80.6   & 90.3      & -       \\
\midrule
\multicolumn{6}{c}{Prompt tuning pre-trained models}               \\
\midrule
\textsc{PTR}                            & w/o  & 72.4   & 81.4   & 90.9     & 89.9      \\
\textsc{KnowPrompt}                     & w/o  & 72.4   & 82.4   & 91.3     & 90.2      \\
\textsc{\textbf{LabelPrompt}}(ours)     & w/o  & \textbf{73.1}   & \textbf{82.5}   & \textbf{91.6}      & \textbf{91.3}      \\
\bottomrule
\end{tabular}
\end{table*}

\subsection{Datasets}
To better verify the effectiveness of our idea, we use the most popular relation classification datasets, including  TACRED~\cite{zhang-etal-2017-tacred}, TACREV~\cite{alt-etal-2020-tacrev}, ReTACRED~\cite{retacred}, and SemEval~\cite{hendrickx-etal-2010-semeval}.

$\textbf{TACRED}$ is one of the most enormous and widely used crowd-sourced datasets in relation extraction. It is created by combining available human annotations from the TAC KBP challenges and crowd-sourcing, and it contains 42 relation types (including ``\textit{no\_relation}'').

$\textbf{TACREV}$ is a modified version of TACRED. It corrects the label errors in the validation and test sets while leaving the training set unchanged.

$\textbf{ReTACRED}$ is a new completely re-annotated version of the TACRED dataset. The dataset fixes many labelling errors in the TACRED dataset and refactors the training, validation, and test sets.

$\textbf{SemEval}$ is a traditional dataset in relation classification. It contains 9 symmetric relations and one special relation ``\textit{Other}''.

\textbf{Annotation quality of the datasets}. In the field of natural language processing, an essential component of dataset preparation is the quality of the annotation. A comparative analysis of the annotation quality of four datasets. The results indicate that the ReTACRED dataset has the highest annotation quality compared to the other three datasets. Although the TACREV and TACRED datasets share the same training data, \citet{alt-etal-2020-tacrev} improved the quality of annotations in the TACREV test set. The statistical details of each dataset have been tabulated in Table~\ref{tab:datasize}.

\subsection{Baselines}

In this paper, we compare the proposed LabelPrompt method with several representative approaches for relation classification. LabelPrompt is a novel method that leverages label prompt tokens to enhance pre-trained language models for this task. We select baselines that cover different aspects of relation classification, such as fine-tuning, prompt-based learning, and knowledge integration.

For fine-tuning pre-trained models, RoBERTa is a very popular language model that improved BERT’s training strategy to overcome its under-training problem, and it also served as the text encoder of this paper.
We also revise some other pre-trained models that incorporated different types of knowledge into their representations.
For example, KnowBERT embeds multiple Knowledge Bases (KBs) into PLM to leverage structured, human-curated knowledge.
MTB builds task-agnostic relation representations from entity-linked text and fine-tuned them on supervised relation extraction datasets.
SpanBERT focuses on improving span representations by defining a masked span prediction task for representing and predicting spans of text with varying lengths.
LUKE predicts randomly masked words and entities in a large entity-annotated corpus retrieved from Wikipedia.

For prompt-based learning models, we select two representative works of prompt-based learning models. PTR~\cite{han2021ptr} uses logic rules to create prompts with sub-prompts that encode the prior knowledge of each class.   KnowPrompt~\cite{chen2022knowprompt} injects latent knowledge in relation labels into the prompts, thus injecting the knowledge into relation labels.

\subsection{Experiment Settings}
We use $\textrm{RoBERTa}_{large}$~\cite{liu2019roberta} as the pre-trained language model for all our experiments and follow the same setting as the previous methods.
We compare the performance of our proposed method with other methods in two scenarios: full-data and few-shot.
In the full-data scenario, we use all the training samples in the dataset for network tuning.
In the few-shot scenario, we use the $k$-shot approach, where $k \in [8, 16, 32]$, to evaluate our model’s performance. The $k$-shot approach means that we sample $k$ instances from each class in the training/validating set as the training/validating data while using all the test instances to evaluate the model’s performance.

We conducted our experiments on NVIDIA GeForce RTX 2080 Ti. We set the batch size to 16 and the learning rate to 4e-5 and 4e-6 for the few-shot and full-data scenarios, respectively.
% More detailed training settings are provided in Appendix~\ref{app:training_details}.
% To obtain more efficient and realistic results in the few-shot scenario, we evaluate the performance of the model under different random seed settings, $seed \in \{ 1, 2, 3, 4, 5 \}$. We consider the average $F_1$ score as the final evaluation metric in this paper.

% \begin{figure}[t]
%     \centering
%     \includegraphics[width = 1\linewidth]{mask-startegy.pdf}
%     \caption{The strategy of mask tokens and unused tokens.}
%     \label{fig:alb-mask}
% \end{figure}

\begin{figure}[t]
    \centering
    \includegraphics[width = 0.6\linewidth]{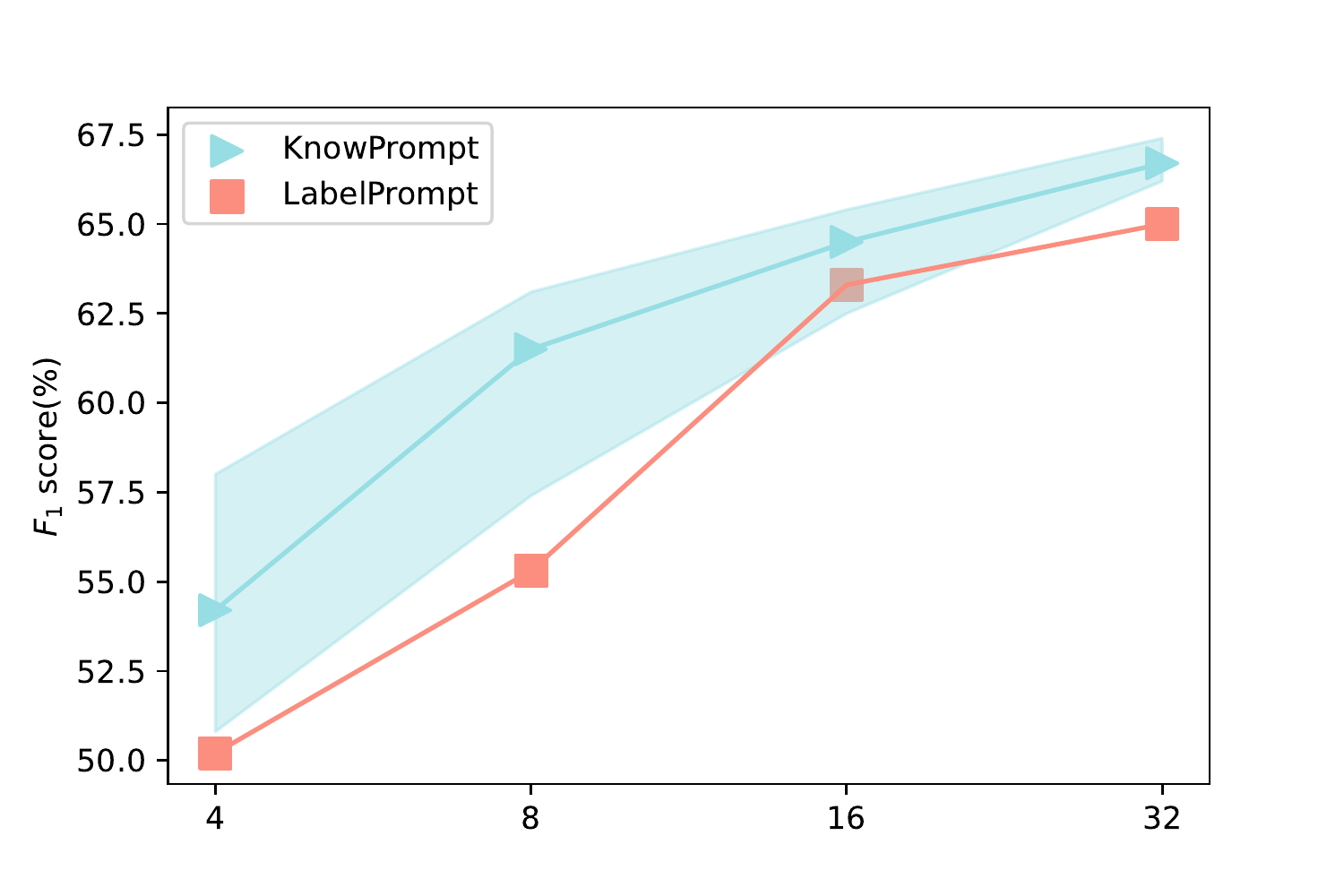}
    \caption{The $F_1$ score of different $k$-shot compared with KnowPrompt.}
    \label{fig:4-shot}
\end{figure}

\subsection{Results}
We first evaluate and analyse the proposed LabelPrompt method by comparing it with both fine-tuning and prompt-based methods. The experimental results are reported in Table~\ref{tab:few-shot} and Table~\ref{tab:all}.

\textbf{Few-Shot}. In Table~\ref{tab:few-shot}, we compare our approach with the traditional fine-tuning and prompt-based methods in the few-shot scenario. Our approach outperforms the other methods in most cases. Furthermore, on the TACREV and ReTACRED datasets, which have more accurate and consistent annotations, our method outperforms all the other methods in all the few-shot settings.

As compared with the fine-tuning methods, our results show that prompt-based methods perform better in the few-shot scenarios for the NLU tasks with low resources. We attribute this to the fact that prompt-based learning effectively uses the knowledge stored in PLMs to fit the task. In contrast, fine-tuning methods rely on complex decoding modules and training strategies based on PLMs, which require sufficient data.

As compared with the previous best prompt-based methods, LabelPrompt achieves 1.5\%-6.2\% performance gains on both the ReTACRED and TACREV datasets, which have high labelling quality. It also achieves 1.9\%-2.7\% improvements on the TACRED and SemEval datasets in the 8-shot setting.

These two aspects demonstrate that the explicit label tokens can aid the pre-trained model in the few-shot scenario. We believe that these additional tokens can reduce the search space for target tokens during the early stage of model training. To further validate our model’s few-shot performance, we design an additional 4-shot setting, as shown in Fig.~\ref{fig:4-shot}. The experimental results demonstrate that, even with only 4 instances per class, the proposed method achieves the same performance as other methods in the 8-shot setting.

\textbf{Full-Data}. In the full-data scenario, as shown in Table~\ref{tab:all}, the proposed method also beat all the other fine-tuning and prompt-based approaches.
% From the experimental results, LabelPrompt outperforms KnowPrompt and other prompt-based methods on all datasets.

In comparison with KnowPrompt, LabelPrompt improves the performance by 0.7\% in TACRED and by 1.1\% in SemEval.
This demonstrates that the prompt-learning method can be applied to both few-shot and full-data tasks. In fact, LabelPrompt leverages the knowledge embedded in PLMs to obtain remarkable results in downstream tasks with a simple design.

So far, we've discussed the performance of the proposed method, and it shows that LabelPrompt is superior to the other methods in both few-shot and full-data scenarios.

\subsection{Ablation Study }

\begin{figure}[t]
    \centering
    \includegraphics[width = 1.0\linewidth]{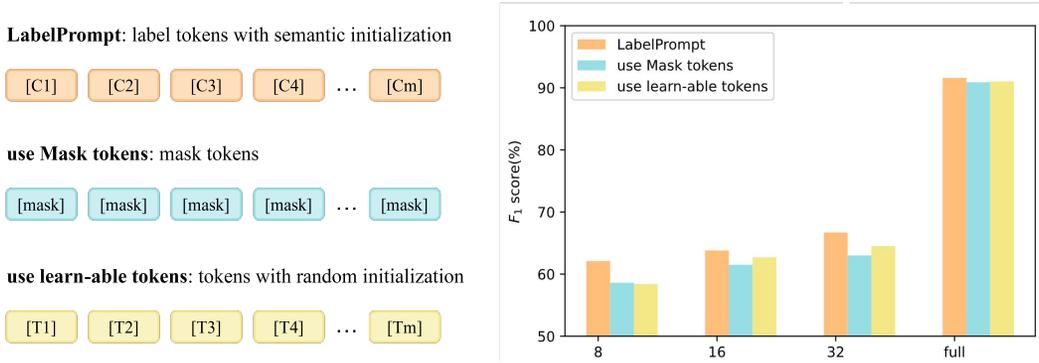}
    \caption{Ablation experiments on different token strategies. \textsc{use Mask tokens} means we replace all label tokens in the prompt template with a special token \texttt{[MASK]}. \textsc{use Learnable tokens} means we replace the label tokens in the prompt template with $m$ additional learnable tokens.}
    \label{fig:ablation-tokens}
\end{figure}

In this section, we evaluate the performance of each component of LabelPrompt in both few-shot and full-data settings. The results are presented in Table~\ref{tab:ablation} and Fig.~\ref{fig:ablation-tokens}.

% \subsubsection{Effect of Label Prompt Tokens}

\begin{table}[t]
\center
\caption{The $F_1$ scores (\%) of ablation experiments in ReTACRED. }
\label{tab:ablation}
\begin{tabular}{l|c|c|c|c}
\toprule
\textbf{Method}      &\textbf{8-shot} & \textbf{16-shot} & \textbf{32-shot} & \textbf{full} \\
\midrule
LabelPrompt                 &\textbf{62.1}  & \textbf{63.8} & \textbf{66.7} & \textbf{91.6}   \\
\midrule
 - w/o label prompt tokens  & 57.6          & 61.6          & 64.7          & 91.5   \\
 - w/o entity-aware module  & 58.9          & 61.7          & 64.7          & 91.2   \\
 - w/o attention query strategy & 61.8      & 62.7          & 65.4          & 90.9   \\
 % - use Mask tokens          & 58.6          & 61.5          & 63.0          & 90.9   \\
 % - use Learnable tokens    & 58.4          & 62.7          & 64.5          & 91.0   \\
\bottomrule
\end{tabular}
\end{table}

% \begin{figure}[t]
%     \centering
%     \includegraphics[width = 1.0\linewidth]{vis_embed_8.pdf}
%     \caption{The visualisation of the hidden vector of test instances in ReTACRED(except for ``no relation'') in two-dimensional space. Different colours indicate different relations.}
%     \label{fig:vis_embeds_paper}
% \end{figure}

\begin{figure*}[!t]
    \centering
    \includegraphics[width = 1\linewidth]{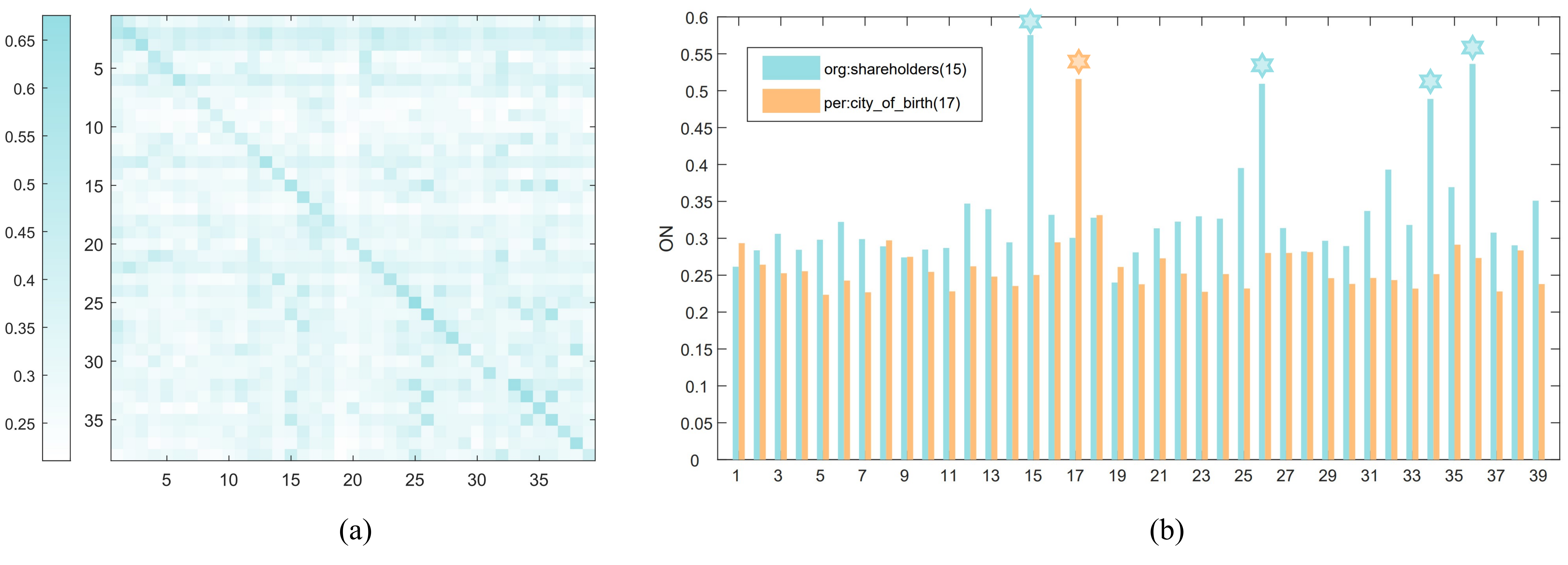}
    \caption{The left subplot shows the \textbf{ON} between the mask token and label tokens for different relations. The right subplot compares the data from rows 15 and 17 of the left subplot in two colours. For example, the blue histogram represents the \textbf{ON} between the mask token and label tokens when the true label is ``\textit{org:shareholders}''.}
    \label{fig:app-vis-acts-all}
\end{figure*}

% \begin{figure*}[!t]
%     \centering
%     \includegraphics[width = 1.0\linewidth]{vis_embed_all-22-11-24.pdf}
%     \caption{The visualisation of the hidden vectors on the 32-shot scenario.}
%     \label{fig:app_vis_embeds_all}
% \end{figure*}

\subsubsection{Impact of Label Prompt Tokens}

The prompt template is one of the most important factors affecting the effectiveness of prompt-based methods. And our prompt template method relies on label prompt tokens, which are very influential. Therefore, we need to demonstrate their impact with more experiments.

First, we ablate the label prompt tokens from the input and remove the label objective module $\mathcal{L}_{label}$ from the overall loss function.
During training, by removing the label prompt tokens, the model must search over the full vocabulary space to predict the correct labels without any explicit label prompt. This significantly increases the difficulty of the prediction task, making it challenging for the model to converge efficiently.
As shown in Table~\ref{tab:ablation}, removing the label prompts leads to a substantial performance degradation, especially under few-shot settings.
To verify that the performance gains are due to the semantic guidance from label prompts rather than simply longer prompt lengths, we evaluate two additional ablated configurations: 1) replacing all label tokens in $T(\mathbf{x})$ with $[\texttt{MASK}]$, and 2) substituting the label tokens with randomly initialised, learnable prompt embeddings.

Fig.~\ref{fig:ablation-tokens} shows that both configurations result in a significant performance drop in few-shot settings. This suggests that label tokens are effective due to their high correlation with the labels in the classification task. Moreover, adding more learnable tokens does not help, as the model cannot fit them with limited training data and performs worse in full-data settings.

To visualise the effectiveness of label prompt tokens, we project the hidden vectors of mask token outputs onto a two-dimensional space based on different relations. The ReTACRED test set samples (excluding "no relation" and "per:country\_of\_birth") are used for visualisation, with a total of 5,648 samples covering 38 different relations.

% Figure~\ref{fig:vis_embeds_paper} shows the visualisation results with and without label prompt tokens, denoted by (a) and (b) respectively. The points representing different relations in (b) are more aggregated, and the boundaries between different relations are clearer.

\subsubsection{Impact of The Entity-Aware Module}

We propose an Entity-Aware Module to improve the performance of the proposed method for relation classification. This module helps the model to focus on the relation class between the entities in a sentence. As shown in Table~\ref{tab:ablation}, removing the Entity-Aware Module decreases the accuracy in all scenarios. The full-data scenario is mostly affected by this removal.

In Table~\ref{tab:case-study}, we provide examples of relation extraction by our model with and without the entity-aware module.  The entity-aware module improves the accuracy of predicting the relations between entities in a sentence, as shown by the results for No.1254 and No.9035. Without the entity-aware module, the model only considers the relations implied in the sentence, while with the entity-aware module, it takes both entities into account for more accurate relation prediction.

\subsubsection{Impact of Attention Query Strategy}

Label prompt tokens are not defined during the pre-training phase of PLM. Although these tokens are initialised with the semantics of the label text, they can still affect the semantics of the entire sentence when the model processes the sentence sequence. As presented in Table 5, when the attention query strategy is removed from our model, we observe varying degrees of performance degradation in all scenarios.

\begin{table}
\center
\caption{Case study for the entity-aware module. The \underline{underlined} text represents the ground-truth relation class.}
\label{tab:case-study}
\begin{tabular}{p{0.65\linewidth}|p{0.25\linewidth}}
\toprule
\textbf{Sentence} & \textbf{Result}  \\
\midrule
\textit{\textbf{[NO.1254]}} It has been a little over a year since Travis the Chimp, the \textit{14-year-old [object]}, 200-pound pet of \textit{Sandra Herold [subject]}, 71, mauled a family friend in Herold's driveway. & \textbf{w/o :} per:age  \newline \textbf{w/ :} \underline{no\_relation} \\
\midrule
\textit{\textbf{[NO.9035]}} \textit{Knox [object]}'s mother Edda Mellas was shaking with grief as \textit{her [subject]} stepmother Cassandra Knox moved to comfort her. & \textbf{w/o :} per:children  \newline \textbf{w/ :} \underline{per:identity} \\
\bottomrule
\end{tabular}
\end{table}

% More details about the visualisation experiments can be found in the Appendix~\ref{app:vis_details}.

% \subsubsection{Effect of Entity-aware Module}

% we designed the special entity-aware module for relation classification methods. to prove the efficiency of this module, we designed t

\section{Analysis}\label{sec:analysis}

\begin{figure}[!t]
    \centering
    \includegraphics[width = 0.6\linewidth]{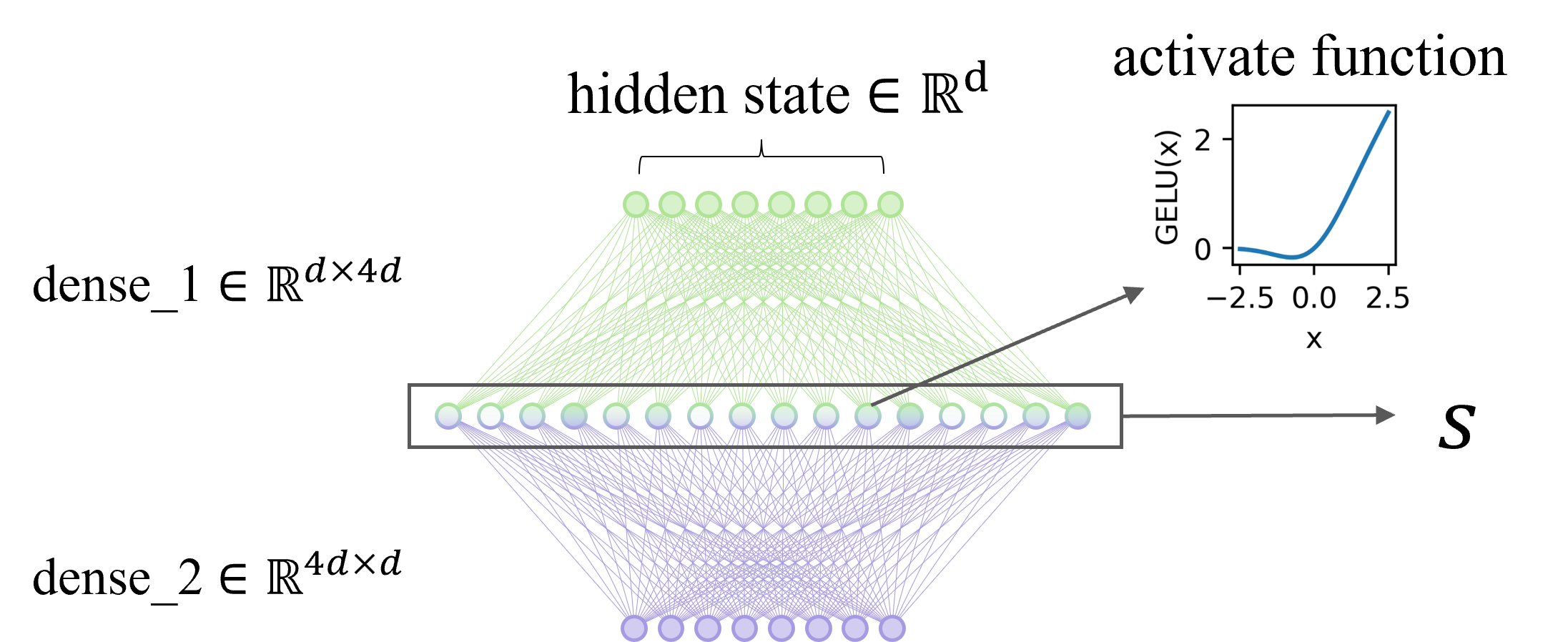}
    \caption{Illustration for activated value sequence $s$.}
    \label{fig:on}
\end{figure}

In this section, we will conduct a detailed analysis of why label prompt tokens are useful for the relation classification task.

Some studies~\cite{geva-etal-2021-transformer, dai-etal-2022-knowledge} regard the Feed-Forward Network (FFN) within the pre-trained language model as a place to store knowledge.
It has been demonstrated that prompt learning is useful for PLMs because prompts can activate specific neurons in the feed-forward network for given inputs. \citet{su-etal-2022-transferability} further considered each node in FFN as a neuron and found that the activation layer of the feed-forward network corresponds to a specific behaviour of the model.

As shown in Fig.~\ref{fig:on}, FFN in a transformer layer can be formulated as:

\begin{equation}
    \mathrm{FFN(\mathbf{x})} = \mathrm{GELU}(\mathrm{\mathbf{x}}\mathbf{W}_{1}^\top + \mathrm{b}_{1})\mathbf{W}_{2} + \mathrm{b}_{2},
\end{equation}
where $\mathrm{\mathbf{x}} \in \mathbb{R}^{d}$ denotes the hidden state of the self-attention output, $\mathrm{\mathbf{W}_1} \in \mathbb{R}^{d \times 4d}$ and $\mathrm{\mathbf{W}_2} \in \mathbb{R}^{4d \times d}$ represent the two dense layers. In this paper, we consider the output of the activation function after the first dense layer in FFN as the activated neurons sequence, denoted as $\mathbf{S}$.

For each instance, the activated neuron sequence can be obtained by:
\begin{equation}
    \mathbf{S} = \{s_{c_1}, s_{c_2}, \dots, s_{c_m}, s_{M}\},
    \label{equ:si}
\end{equation}
where $s_M$ is the activated neuron sequence of mask tokens.

We defined a function $\textbf{ON}(s_{a}, s_{b})$ to calculate the \underline{\textbf{O}}verlapping rate of activated \underline{\textbf{N}}eurons (\textbf{ON}) of two activated neuron sequences $s_{a}$ and $s_{b}$:
\begin{equation}
    \mathrm{\mathbf{ON}}(s_{a}, s_{b}) = \frac{\sum_{k=1}^{4d} \zeta(s_{a,k} > 0 \wedge s_{b, k} > 0)}{\sum_{k=1}^{4d} \zeta(s_{a,k} > 0) + \sum_{k=1}^{4d} \zeta(s_{b,k} > 0)},
\end{equation}
where $\zeta(c)$ stands for a binary function.
The result is 1 if the condition $c$ is true and vice versa.

The \textbf{ON} function between different tokens can be used as the associated probability of the different corresponding behaviours of the model. Therefore, we compare \textbf{ON} between label prompt tokens $\mathcal{C}$ and the mask token \texttt{[MASK]} at the last encoder layer. We then take \textbf{ON} as the task-driven capability of label prompt tokens.

The data used for visualisation and analysis contains 5648 samples and 38 relations (excluding ``\textit{no relation}'' and ``\textit{per:country\_of\_birth}'') from ReTACRED.
To obtain Fig.~\ref{fig:app-vis-acts-all}, we record the activated values in the last encoder layer for all the test instances. We consider the neurons with positive values as activated neurons.
To avoid the distribution error caused by sample sampling, we then average the overlapping rate of all samples of relation $i$ as the $\mathrm{\mathbf{ON}}$.

As shown in Fig.~\ref{fig:app-vis-acts-all}~(a), each point $r(i,j)$ is defined as:
\begin{equation}
\begin{aligned}
    r(i,j) = \frac{1}{|\mathcal{C}_{i}|} \sum_{u=1}^{|\mathcal{C}_{i}|} \mathrm{\mathbf{ON}}(s_j^{(u)}, s_M^{(u)}) \hspace{2em} \mathrm{s.t.} \hspace{1em} gt = c_i,
\end{aligned}
\label{equ:on-simple}
\end{equation}
which denotes the \textbf{ON} result between the mask token and label token $c_j$ when the ground-truth relation is $c_i$.
Fig.~\ref{fig:app-vis-acts-all}~(b) represents $\mathbf{ON}$ of the relations ``\textit{org:shareholders}'' and ``\textit{per:city\_of\_birth}''.

According to Fig.~\ref{fig:app-vis-acts-all}~(a), the colour of the diagonal blocks is darker. The block on the diagonal $r(i,i)$ represents the \textbf{ON} results between the mask token sequence $s_M$ and the ground-truth label prompt token sequence $s_{c_i}$.
\textit{i.e.} the mask token always has a higher overlapping rate with the label token corresponding to the ground-truth label.
This indicates that the label token can lead the mask token to activate the knowledge of the corresponding relation stored in a PLM.

Furthermore, we choose two representative relations to illustrate the results. Fig.~\ref{fig:app-vis-acts-all}~(b) shows the histograms with two rows of data from Fig.~\ref{fig:app-vis-acts-all}~(a).
The histograms clearly show the difference in the overlapping rate between the mask and label tokens. The histogram in orange is the result of the relation ``\textit{org:shareholders}'', compared to other label tokens, the \textbf{ON} result between mask token and label token 17 is at the peak.
Meanwhile, the histogram in blue shows that when the relation is ``\textit{per:city\_of\_birth}'', the overlapping rate is high not only for token 15, but also for tokens 26, 34, 36, and so on. In fact, all of these tokens also have ``\textit{city}'' in their labels, which is consistent with our intuition that these relations have a high overlapping ratio of activated knowledge of their relations.

\section{Conclusion}
\label{sec:conclusion}
In this paper, we proposed LabelPrompt, a strategy to effectively apply label information for the relation classification task based on prompt learning and prompt engineering.
By explicitly adding label tokens to sentences before feeding them into the model, the model effectively adapted downstream tasks with our LabelPrompt.
The additional entity-aware module also alleviated the problem of relations in sentences that do not correspond to entities.

The experimental results demonstrated that the label prompt tokens are effective in both the few-shot and full-data scenarios. The method focused more on the additional label tokens and correctly predicted the relation class of the entities in a sentence.
In future work, we plan to improve our method's performance on poorly annotated datasets in low-resource scenarios and extend our work to other text classification tasks.

\begin{acks}
This work was supported in part by Major Project of the National Social Science Foundation of China (No. 21\&ZD166), National Natural Science Foundation of China (61876072) and Natural Science Foundation of Jiangsu Province (No. BK20221535).
\end{acks}

\bibliographystyle{ACM-Reference-Format}
\bibliography{sample-base}

\end{document}